\documentclass{article} 

\usepackage{graphicx}
\usepackage{amsmath}
\usepackage{amssymb}
\usepackage{amsthm}
\usepackage{mathrsfs}
\usepackage{psfrag}
\usepackage{booktabs}
\usepackage[square, comma, sort&compress, numbers]{natbib}
\usepackage{nips11submit_e,times}

\newcommand{\comment}[1]{}

\newcommand{\prob}{\mathsf{P}}      

\theoremstyle{plain}

\theoremstyle{definition}

\title{Bayesian Causal Induction}

\author{
Pedro A. Ortega \\
Max-Planck Institute for Biological Cybernetics\\
Spemannstra{\ss}e 38, 72070 T\"{u}bingen \\
\texttt{peortega@dcc.uchile.cl} \\
}

\nipsfinalcopy 

\begin{document}

\maketitle

\begin{abstract}
Discovering causal relationships is a hard task, often hindered by the need for intervention, and often requiring large amounts of data to resolve statistical uncertainty. However, humans quickly arrive at useful causal relationships. One possible reason is that humans extrapolate from past experience to new, unseen situations: that is, they encode beliefs over causal invariances, allowing for sound generalization from the observations they obtain from directly acting in the world.

Here we outline a Bayesian model of causal induction where beliefs over competing causal hypotheses are modeled using probability trees. Based on this model, we illustrate why, in the general case, we need interventions plus constraints on our causal hypotheses in order to extract causal information from our experience.
\end{abstract}

\section{Introduction}

A fundamental problem of statistical causality is the problem of \emph{causal induction}\footnote{For a thorough treatment of non-causal induction, refer to \citep{Rathmanner2011}.}; namely, the generalization from particular instances to abstract causal laws \citep{Hume1739, Griffiths2007}. For instance, how can you conclude that it is dangerous to ride a bike on ice from a bad slip fall on wet floor?

In this work, we are concerned with the following problem: how do we determine from experience whether ``$X \rightarrow Y$ and $U \rightarrow V$'' or ``$Y \rightarrow X$ and $V \rightarrow U$''? That is, which of the two causal hypotheses over $X$, $Y$, $U$ and $V$ is correct,
\begin{center}
    \small
    \psfrag{x1}[c]{$X$}
    \psfrag{x2}[c]{$Y$}
    \psfrag{x3}[c]{$U$}
    \psfrag{x4}[c]{$V$}
    \psfrag{a1}[c]{or}
    \psfrag{a2}[c]{,}
    \includegraphics{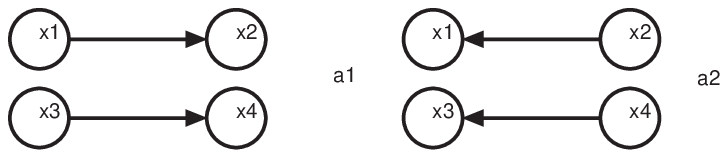}
\end{center}
even in the case when both models represent identical joint distributions? Furthermore, if we collect evidence supporting the claim ``$X \rightarrow Y$'', how do we extrapolate this to the (yet unseen) situation ``$U \rightarrow V$''? The main challenge in this problem is that the hypothesis, say $H$, is a random variable that controls the very causal structure. That is, a more accurate graphical representation would be the model:
\begin{center}
    \small
    \psfrag{x1}[c]{$X$}
    \psfrag{x2}[c]{$Y$}
    \psfrag{x3}[c]{$U$}
    \psfrag{x4}[c]{$V$}
    \psfrag{x5}[c]{$H$}
    \psfrag{a1}[c]{meta-level}
    \includegraphics{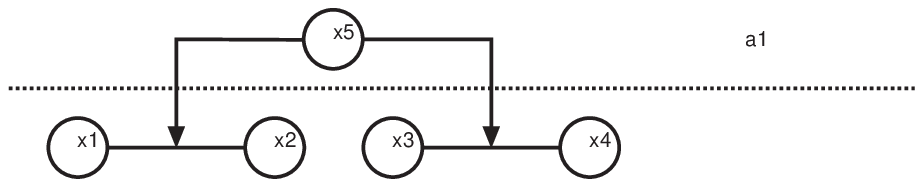}
\end{center}
which cannot be analyzed using the framework of graphical models alone because the random variable $H$ operates on a meta-level of the graphical model over $X, Y, U$ and $V$.

In this work these difficulties are overcome by using a probability tree to model the causal structure over the random events \citep{Shafer1996}. Probability trees can encode\footnote{Conditional independencies are also captured within a probability tree \citep[Chapter 8.2]{Shafer1996}.} alternative causal realizations, and in particular alternative causal hypotheses. All random variables are of the same type---no distinctions between meta-levels are needed. Furthermore, we define interventions \citep{Pearl2009} on probability trees so as to predict the statistical behavior after manipulation. We then show that such a formalization leads to a probabilistic method for causal induction that is intuitively appealing.

\section{Causal Induction in Probability Trees}

\begin{figure}[tbp]
\begin{center}
    \scriptsize
    \psfrag{x1}[c]{$X$}
    \psfrag{x2}[c]{$Y$}
    \psfrag{x3}[c]{$U$}
    \psfrag{x4}[c]{$V$}
    \includegraphics{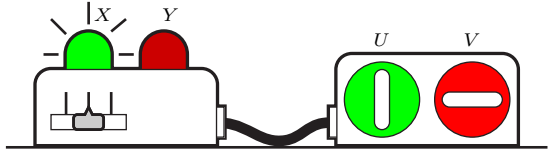}
    \caption{(Left) A device with a green and a red light bulb. A switch allows controlling the state of the green light: either ``on'', ``off'' or ``undisturbed''. (Right) A second device having a green spinner and a red spinner, both of which can either lock into a horizontal or vertical position. The two devices are connected through a cable, establishing thus a relation among their randomizing mechanisms.}
    \label{fig:device}
\end{center}
\end{figure}

Imagine we are given a device with two light bulbs, one green ($X$) and one red ($Y$), whose states obey a hidden mechanism that correlates them positively. Moreover, the box has a switch that allows us controlling the state of the green bulb: we can either leave it undisturbed, or we can intercept the mechanism by turning the light on or off as we please (Figure~\ref{fig:device}, left device). We encode the ``on'' and ``off'' states of the green light as $X=x$ and $X=\neg x$ respectively. Analogously, $Y=y$ and $Y=\neg y$ denote the ``on'' and ``off'' states of the red light. We ponder the explanatory power of two competing hypotheses: either ``green causes red'' ($H=h$) or ``red causes green'' ($H=\neg h$).

\subsection{Representation}

\begin{figure}[tbp]
\begin{center}
    \scriptsize
    \psfrag{l1}[l]{a)}
    \psfrag{l2}[l]{b)}
    \psfrag{p00}[c]{$\frac{1}{2}$}
    \psfrag{p01}[c]{$\frac{1}{2}$}
    \psfrag{p02}[c]{$\frac{1}{2}$}
    \psfrag{p03}[c]{$\frac{1}{2}$}
    \psfrag{p04}[c]{$\frac{1}{2}$}
    \psfrag{p05}[c]{$\frac{1}{2}$}
    \psfrag{p06}[c]{$\frac{3}{4}$}
    \psfrag{p07}[c]{$\frac{1}{4}$}
    \psfrag{p08}[c]{$\frac{1}{4}$}
    \psfrag{p09}[c]{$\frac{3}{4}$}
    \psfrag{p10}[c]{$\frac{3}{4}$}
    \psfrag{p22}[c]{$\frac{1}{4}$}
    \psfrag{p12}[c]{$\frac{1}{4}$}
    \psfrag{p13}[c]{$\frac{3}{4}$}
    \psfrag{p14}[c]{$\frac{3}{16}$}
    \psfrag{p15}[c]{$\frac{1}{16}$}
    \psfrag{p16}[c]{$\frac{1}{16}$}
    \psfrag{p17}[c]{$\frac{3}{16}$}
    \psfrag{p18}[c]{$\frac{3}{16}$}
    \psfrag{p19}[c]{$\frac{1}{16}$}
    \psfrag{p20}[c]{$\frac{1}{16}$}
    \psfrag{p21}[c]{$\frac{3}{16}$}
    \psfrag{a00}[c]{$\Omega$}
    \psfrag{a01}[c]{$h$}
    \psfrag{a02}[c]{$\neg h$}
    \psfrag{a03}[c]{$x$}
    \psfrag{a04}[c]{$\neg x$}
    \psfrag{a05}[c]{$y$}
    \psfrag{a06}[c]{$\neg y$}
    \psfrag{a07}[c]{$y$}
    \psfrag{a08}[c]{$\neg y$}
    \psfrag{a09}[c]{$y$}
    \psfrag{a10}[c]{$\neg y$}
    \psfrag{aaa}[c]{$x$}
    \psfrag{a12}[c]{$\neg x$}
    \psfrag{a13}[c]{$x$}
    \psfrag{a14}[c]{$\neg x$}
    \psfrag{q00}[c]{$\frac{1}{2}$}
    \psfrag{q01}[c]{$\frac{1}{2}$}
    \psfrag{q02}[c]{$\bf{1}$}
    \psfrag{q03}[c]{$\bf{0}$}
    \psfrag{q04}[c]{$\frac{1}{2}$}
    \psfrag{q05}[c]{$\frac{1}{2}$}
    \psfrag{q06}[c]{$\frac{3}{4}$}
    \psfrag{q07}[c]{$\frac{1}{4}$}
    \psfrag{q08}[c]{$\frac{1}{4}$}
    \psfrag{q09}[c]{$\frac{3}{4}$}
    \psfrag{q10}[c]{$\bf{1}$}
    \psfrag{q22}[c]{$\bf{0}$}
    \psfrag{q12}[c]{$\bf{1}$}
    \psfrag{q13}[c]{$\bf{0}$}
    \psfrag{q14}[c]{$\frac{3}{8}$}
    \psfrag{q15}[c]{$\frac{1}{8}$}
    \psfrag{q16}[c]{$0$}
    \psfrag{q17}[c]{$0$}
    \psfrag{q18}[c]{$\frac{1}{4}$}
    \psfrag{q19}[c]{$0$}
    \psfrag{q20}[c]{$\frac{1}{4}$}
    \psfrag{q21}[c]{$0$}
    \psfrag{b00}[c]{$\Omega$}
    \psfrag{b01}[c]{$h$}
    \psfrag{b02}[c]{$\neg h$}
    \psfrag{b03}[c]{$x$}
    \psfrag{b04}[c]{$\neg x$}
    \psfrag{b05}[c]{$y$}
    \psfrag{b06}[c]{$\neg y$}
    \psfrag{b07}[c]{$y$}
    \psfrag{b08}[c]{$\neg y$}
    \psfrag{b09}[c]{$y$}
    \psfrag{b10}[c]{$\neg y$}
    \psfrag{baa}[c]{$x$}
    \psfrag{b12}[c]{$\neg x$}
    \psfrag{b13}[c]{$x$}
    \psfrag{b14}[c]{$\neg x$}
    \includegraphics{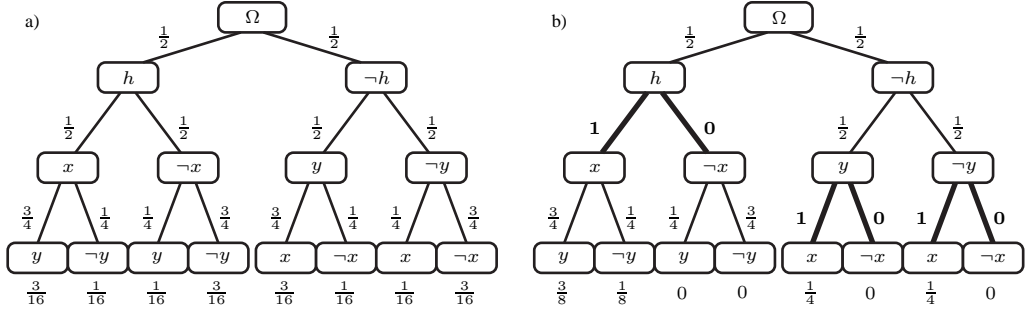}
    \caption{a) The probability tree representing the statistics of the device with two lights. The probability of a realization (written under the leaves) is calculated by multiplying the probabilities starting from the root until a leave is reached. Note that the two hypotheses are statistically indistinguishable. b) The probability tree resulting from (a) after setting $X=x$.}
    \label{fig:tree}
\end{center}
\end{figure}

Assume that the probabilities governing the realization of $H$, $X$ and $Y$ are as detailed in Figure~\ref{fig:tree}a. In this tree, each (internal) node is interpreted as a causal mechanism; hence a path from the root node to one of the leaves corresponds to a particular sequential realization of causal mechanisms\footnote{Note that the set of paths is the sample space of the experiment's probability space.}. The logic underlying the structure of this tree is self-explanatory:
\begin{enumerate}
    \item \emph{Causal precedence:} A node causally precedes its descendants. For instance, the root node corresponding to the sure event $\Omega$ causally precedes all other nodes.
    \item \emph{Resolution of variables:} Each node resolves the value of a random variable. For instance, given the node corresponding to $H=h$ and $X=\neg x$, either $Y=y$ will happen with probability $\prob(y|h,\neg x) = \frac{1}{4}$ or $Y=\neg y$ with probability $\prob(\neg y|h,\neg x) = \frac{3}{4}$.
    \item \emph{Heterogeneous order:} The resolution order of random variables can vary across different branches. For instance, $X$ precedes $Y$ under $H=h$, but $Y$ precedes $X$ under $H=\neg h$. This allows modeling different causal hypotheses.
\end{enumerate}
While the probability tree represents our subjective model explaining the order in which the random values are resolved, it does not necessarily correspond to the temporal order in which the events are revealed to us. So for instance, under hypothesis $H=h$, the value of the variable $Y$ might be revealed before $X$, even though $X$ causally precedes $Y$; and the hypothesis $H$, which precedes both $X$ and $Y$, is never observed.

\subsection{Interventions}

Suppose we observe that both lights are on. Have we learned anything about their causal dependency? A brief calculation shows that this is not the case because the posterior probabilities are equal to the prior probabilities:
\small
\begin{align*}
    \prob(h|x,y)
        &= \frac{\prob(y|h,x)\prob(x|h)\prob(h)}
                {\prob(y|h,x)\prob(x|h)\prob(h)
                + \prob(x|\neg h,y)\prob(y|\neg h)\prob(\neg h)}
        = \frac{\frac{3}{4} \cdot \frac{1}{2} \cdot \frac{1}{2}}
               {\frac{3}{4} \cdot \frac{1}{2} \cdot \frac{1}{2}
               + \frac{3}{4} \cdot \frac{1}{2} \cdot \frac{1}{2}}
        = \frac{1}{2} = \prob(h).
\end{align*}
\normalsize
This makes sense intuitively, because by just observing that the two lights are on, it is statistically impossible to tell which one caused the other. Note how the factorization of the likelihood $P(x,y|H)$ depends on whether $H=h$ or $H=\neg h$. How do we extract causal information then? To answer this question, we make use of a crucial insight of statistical causality:
\begin{quote}
To obtain new causal information from statistical data, old causal information needs to be supplied (paraphrased as ``no causes in, no causes out'' \citep{Cartwright1994} or ``to find out what happens when you kick the system, you have to kick the system'' \citep{Box1966}).
\end{quote}
Thus, we now repeat our experiment, but this time \emph{we} turn on the green light ($X=x$). We reflect this choice by changing all the mechanisms that resolve the random variable $X$, placing all the probability mass on the outcome $X=x$ (see Figure~\ref{fig:tree}b). Assume that we subsequently observe that the second light is on. Then, the posterior probabilities are
\small
\begin{align*}
    \prob(h|\hat{x},y)
        &= \frac{\prob(y|h,\hat{x})\prob(\hat{x}|h)\prob(h)}
                {\prob(y|h,\hat{x})\prob(\hat{x}|h)\prob(h)
                + \prob(\hat{x}|\neg h,y)\prob(y|\neg h)\prob(\neg h)}
        = \frac{\frac{3}{4} \cdot 1 \cdot \frac{1}{2}}
               {\frac{3}{4} \cdot 1 \cdot \frac{1}{2}
                + 1 \cdot \frac{1}{2} \cdot \frac{1}{2}}
        = \frac{3}{5},
\end{align*}
\normalsize
where $\hat{x}$ is Pearl's notation to indicate a causal intervention of $X$. Since $P(h) > P(h|\hat{x},y)$, we have gathered evidence favoring the hypothesis ``green causes red''. This was only possible because our intervention introduced a statistical asymmetry among the two hypotheses that did not exist before.

\subsection{Extrapolation}

Let us now connect a second device to the first one (Figure~\ref{fig:device}, right device). This device carries two spinners, a green ($U$) and a red one ($V$). A hidden randomizing mechanism chooses their orientations (either horizontal or vertical) independently from the state of the colored lights. However, the connection and the mysterious color coding suggest that there must be a relation between the two randomizing mechanisms. Hence, we impose that the combined system either follows the law ``green causes red'' or ``red causes green''---intentionally excluding the cases ``$X \rightarrow Y$ and $U \leftarrow V$'' and ``$X \leftarrow Y$ and $U \rightarrow V$''.

The probability tree over the random variables $X, Y, U$ and $V$ extends the probability tree from Figure~\ref{fig:tree}a by appending sub-trees over $U$ and $V$ having the restriction that the nodes resolving $U$ precede the nodes resolving $V$ under hypothesis $H = h$, and that the nodes resolving $V$ precede the nodes resolving $U$ in the case $H = \neg h$.

Note however, that for this new tree, the posterior probability over the hypothesis ``green causes red'' given that we turned on the green light and saw the red bulb lighting up is identical to the previous tree, namely $\prob(h|\hat{x},y) = \frac{3}{5}$. The restriction we have imposed over the possible causal hypotheses has enabled us extrapolating causal information from our experience with $X$ and $Y$ to the yet unobserved variables $U$ and $V$. This extrapolation would not have been possible if we had kept all four causal hypotheses. Hence, in the general case, causal extrapolation rests on constraints on our causal hypotheses.

\section{Concluding Remarks}

The problem of causal induction has been addressed relatively recently by the statistics and machine learning community, mainly under the context of graphical models \citep{Pearl2009, Spirtes2001, Silva2005, Dawid2007, Griffiths2007, Mooij2010a}. This has led to the development of many algorithms that propose a suitable causal graphical model explaining the data. Many of these algorithms rely on independence assumptions, and hence naturally they proceed by exploiting the independence relations found in the data to construct a causal model.

This work outlines a general method for causal induction that is Bayesian in nature and does not rely on independence assumptions. It is based on the idea of combining probability trees \citep{Shafer1996} with interventions \citep{Pearl2009} for predicting the behavior of a manipulated system with multiple causal hypotheses. We have seen that both the interventions and the (constraints on the) causal hypotheses introduce statistical asymmetries that permit the extraction and extrapolation of causal information. Of course, this means that the amount and the forms of causal relations that we can discover are determined (a)~by our constraints on the set of causal hypotheses and (b)~by the interventions that we are allowed to apply to the system (and essentially, to our hypotheses). In a sense, one could say that we are ``imprinting our own causal laws onto our experience''. However, this raises more fundamental questions that we have not addressed here: where do these constraints on our causal hypotheses come from and what logic do they obey?

\subsubsection*{Acknowledgments}

The author would like to thank Daniel A. Braun, Philip Dawid, David Balduzzi, Samory Kpotufe, Theofanis Karaletsos, Eleni Sgouritsa and Marcus Hutter for discussions of previous versions of this manuscript. This study was supported by the Emmy Noether Grant BR 4164/1-1, ``Computational and Biological Principles of Sensorimotor Learning''.

\bibliographystyle{plain}
\small
\bibliography{bibliography}

\end{document}